\documentclass[11pt,a4paper]{article}
\usepackage[hyperref]{emnlp2020}
\usepackage{times}
\usepackage{latexsym}
\usepackage{xspace}
\usepackage{graphicx}
\usepackage{amsfonts}
\usepackage{amssymb}
\usepackage{amsthm}
\usepackage{amsmath}
\usepackage{booktabs}

\usepackage{paralist}

\usepackage{xargs}                      % Use more than one optional parameter in a new commands
\usepackage[colorinlistoftodos,prependcaption,textsize=tiny]{todonotes}
\newcommandx{\ainfo}[2][1=]{\todo[linecolor=blue,backgroundcolor=blue!25,bordercolor=blue,#1]{#2}} 
 
\usepackage{microtype}

\aclfinalcopy  %Uncomment this line for the final submission
\def\aclpaperid{838}   %Enter the acl Paper ID here

\definecolor{matchcolor}{RGB}{194,123,160}
\definecolor{subscolor}{RGB}{103,78,167}
\definecolor{demotecolor}{RGB}{0,0,255}
\definecolor{biascolor}{RGB}{255,0,0}

\newcommand{\Sref}[1]{\S\ref{#1}}

\newcommand{\Fref}[1]{Figure~\ref{#1}}

\newcommand{\Tref}[1]{Table~\ref{#1}}
\newcommand{\Aref}[1]{Appendix~\ref{#1}}

\newcommand\OP{{\textsc{ow}}\xspace}
\newcommand\OPTEXT{{\textsc{o\_txt}}\xspace}
\newcommand\OPGENDER{{\textsc{w\_gen}}\xspace}
\newcommand\OPTRAITS{{\textsc{w\_traits}}\xspace}
\newcommand\COMMENTTEXT{{\textsc{com\_txt}}\xspace}

\newcommand{\ignore}[1]{}

\title{Unsupervised Discovery of Implicit Gender Bias}
\author{Anjalie Field \\
  Carnegie Mellon University \\
  \texttt{anjalief@cs.cmu.edu} \\\And
  Yulia Tsvetkov \\
  Carnegie Mellon University \\
  \texttt{ytsvetko@cs.cmu.edu} \\}

\date{}

\begin{document}
\maketitle
\begin{abstract}
Despite their prevalence in society, social biases are difficult to identify, primarily because human judgements in this domain can be unreliable. We take an \textit{unsupervised} approach to identifying gender bias against women at a comment level and present a model that can surface text likely to contain bias. Our main challenge is forcing the model to focus on signs of implicit bias, rather than other artifacts in the data. Thus, our methodology involves reducing the influence of confounds through propensity matching and adversarial learning.
Our analysis shows how biased comments directed towards female politicians contain mixed criticisms, while comments directed towards other female public figures focus on appearance and sexualization. Ultimately, our work offers a way to capture subtle biases in various domains without relying on subjective human judgements.\footnote{Code and pre-trained models are available at \url{https://github.com/anjalief/unsupervised_gender_bias}}
\end{abstract}

\section{Introduction}
\label{sec:intro}

Despite widespread documentation of the negative impacts of bias, stereotypes, and prejudice \cite{krieger1990racial,goldin1992understanding,steele1995stereotype,Logel2009,schluter-2018-glass}, these concepts remain difficult to define and identify, especially for non-experts. Social biases appear to be a natural component of human cognition that allow people to make judgments efficiently \citep{kahneman1982judgment}. As a result, they are often \emph{implicit}---people are unaware of their own biases \citep{Blair2002,bargh1999cognitive}---and manifest subtly, e.g., as microaggressions or condescension \citep{huckin2002critical,sue2010microaggressions}.

Much NLP literature has examined biases in data, algorithms, or model performance, and the negative pipeline between them: models absorb and amplify data biases, which impacts performance \citep{sun-etal-2019-mitigating}. However, little work has looked further up the pipeline and relied on the assumption that biases in data originate in human cognition. 

In contrast, this assumption motivates our work: an unsupervised approach to detecting implicit gender bias in text. Text provides an ideal avenue for studying bias, because human cognition is closely tied to natural language. Psychology studies often examine human perceptions through word associations \citep{greenwald1998measuring}. However, the implicit nature of bias suggests that human annotations for bias detection may not be reliable, which motivates an unsupervised approach.

The goals of our work align with prior work in NLP that has examined biases in real-world data. However, prior work examines bias at a broad corpus level or relies on supervised models. While corpus-level analyses, e.g. associations between gendered words and stereotypes, can be insightful \citep{bolukbasi2016man,fast2016shirtless,caliskan,garg2018word,friedman-etal-2019-relating,chaloner-maldonado-2019-measuring}, they are difficult to interpret over short text spans. They also often rely on human-defined ``known'' stereotypes, such as lists of traditionally male and female occupations obtained through crowd-sourcing, which restricts analysis to a narrow surface-level domain.  
Similarly, supervised approaches can provide insight into carefully defined types of bias \citep{wang-potts-2019-talkdown,breitfeller2019finding,sap2019social}, but they rely on human annotations tasks, which are difficult to design or generalize to other domains, especially because social concepts differ across contexts and cultures \citep{dong2019perceptions}.

Our work offers a new approach to surfacing gender bias that does not require direct supervision and is meaningful at a sentence or paragraph level.
%Our primary methodology involves unsupervisedly training a model to identify differences in text addressed towards men and women, thereby surfacing gender-biased comments. Machine learning models excel at learning patterns from data, so much so that removing stereotypes and bias from models has become a substantial component of NLP research \citep{bolukbasi2016man,webster-etal-2018-mind, rudinger-etal-2018-gender, zhao-etal-2018-gender,stanovsky-etal-2019-evaluating}. Rather than trying to mitigate this property of machine learning models, we instead leverage it in order to surface bias.
We create a model that takes text in the  2\textsuperscript{nd}-person perspective as input and predicts the gender of the person the text is addressed to. If the classifier predicts the gender of the addressee with high confidence based only on the text directed to them, we hypothesize that the text is likely to contain bias. The main challenge is encouraging the model to focus on text features that are indicative of bias, rather than artifacts in data that correlate with the gender of the addressee but occur because of confounding variables (\textit{confounds}). Thus, the core of our methodology focuses on reducing the influence of confounds. Our goal is not to improve accuracy of the gender-prediction task, but rather to validate that our methodology demotes confounds and surfaces comments likely to contain gender bias.

In \Sref{sec:problem}, we define the problem and intuition behind our approach. We describe our methods for confound demotion in \Sref{sec:methodology}, and we evaluate them in \Sref{sec:evaluation}. Our evaluation involves examining how confound control affects performance on in-domain and out-of-domain classification tasks, including detection of gender-based microaggressions.
Our results suggest that our model successfully identifies text likely to contain bias against women, allowing us to analyze how this bias differs across domains (\Sref{sec:analysis}). To the best of our knowledge, this is the first work that aims to analyze bias in short text spans by learning implicit associations from data sets. 

\section{Problem Formulation}
\label{sec:problem}

\begin{figure}
    \centering
    \includegraphics[width=\columnwidth]{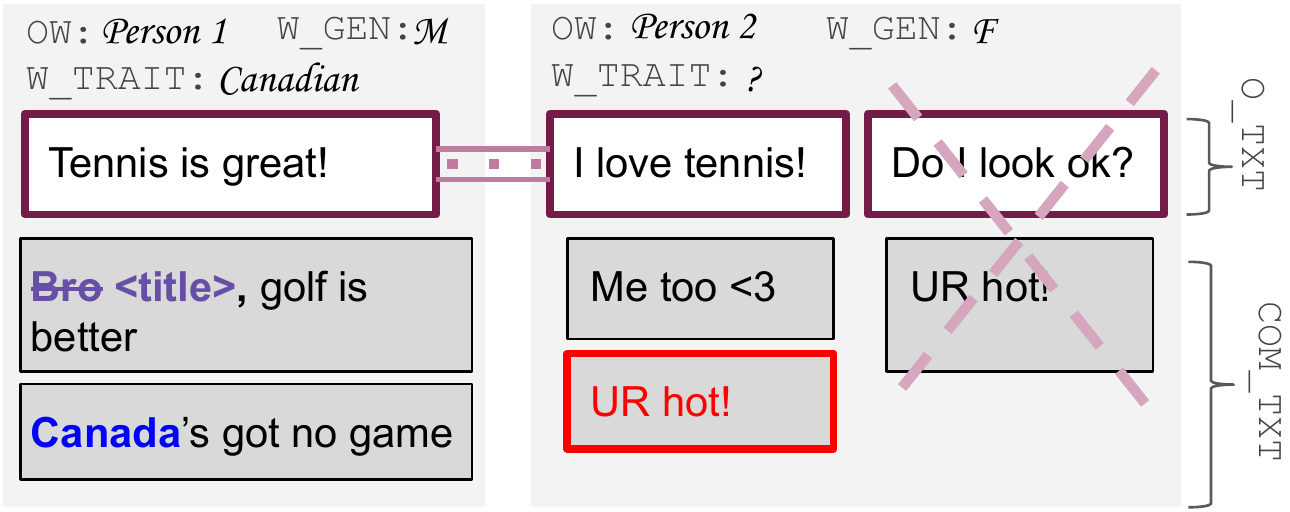}
   \caption{We train a classifier to predict the gender (\OPGENDER) of the person that text is addressed to (\COMMENTTEXT), while demoting features that are predictive of gender but not predictive of bias. Posts with similar content are \textcolor{matchcolor}{matched} through propensity scores; unmatched posts are discarded. Latent traits of the addressee (e.g., nationality) are \textcolor{demotecolor}{demoted} through an adversarial objective. Overtly gendered language (``Bro'') is \textcolor{subscolor}{substituted}. Comments indicative of gender despite these restrictions are likely to contain \textcolor{biascolor}{bias}.}
    \label{fig:method}
\end{figure}

Our primary task is to detect gender bias in a communicative domain, specifically in texts targeting an addressee (i.e., 2\textsuperscript{nd}-person) without relying on explicit bias annotations. Our goals align with a causality framework in that we seek to identify content that occurs because of the gender of the addressee rather than other factors. We can define a counterfactual: \textit{Would the addressee have received different text if their gender were different?}

While our framework is broadly applicable, in order to define consistent notation, we consider a setup where our primary text is a comment written in reply to text written by someone else. This includes domains like replies on social media posts, or comments on newspaper articles, and can be generalized other media, e.g., comments on YouTube videos. We identify the following variables:

\begin{compactitem}
    \item \textbf{\OP}: ``Original Writer'', the person who wrote the original text, e.g., the addressee
    \item \textbf{\OPTEXT}: content of the original text
    \item \textbf{\OPGENDER}: the gender ($\mathbf{M}$, $\mathbf{F}$) of \OP. We use a binary variable because all of the individuals in our corpus identify as men or women, but our methodology is generalizeable and can be used to examine bias against other genders.
    \item \textbf{\OPTRAITS}: any traits of \OP other than gender, e.g., social role, age, nationality.
    % We specifically avoid enumerating these traits.
    \item \textbf{\COMMENTTEXT}: comments replying  to \OPTEXT
\end{compactitem}

Our goal is to detect bias in \COMMENTTEXT values that occurs because of \OPGENDER. A naive approach would train a classifier to predict \OPGENDER from \COMMENTTEXT and assume that any \COMMENTTEXT values for which the classifier correctly predicts \OPGENDER with high confidence contain bias. However, \COMMENTTEXT may contain features that are predictive of \OPGENDER but are not indicative of bias.
 
For example, in \Fref{fig:method}, when the comment ``UR hot!'' (\COMMENTTEXT) is addressed to someone who said ``I love tennis!'' (\OPTEXT), it is an objectification and unsolicited reference to appearance, which could indicate bias. However, when it is addressed to someone who said ``Do I look ok?'', it is likely not indicative of bias. If women ask ``Do I look ok?'' more frequently than men, this naive classifier would identify ``UR hot!'' is likely addressed towards a woman and identify it as biased. However, we only want the model to learn that references to appearance are indicative of gender if they occur in unsolicited contexts. Thus our model needs to account for the effects of \OPTEXT: Because of correlations between \OPGENDER and \OPTEXT, \COMMENTTEXT values may contain features that are \textit{predictive} of \OPGENDER, but are \textit{caused} by \OPTEXT, rather than by \OPGENDER. We face a similar problem with \OPTRAITS. From the synthetic example in \Fref{fig:method}, if our data set contains more men from Canada than women, the model might learn that references to Canada indicate $\OPGENDER=\mathbf{M}$. We provide additional empirical examples in \Sref{sec:experimental}. 

We refer to factors that might influence \COMMENTTEXT as \emph{confounding variables} and the artifacts that they produce in \COMMENTTEXT as \textit{confounds}. We distinguish two types: \textit{observed} and \emph{latent}. Latent confounding variables cannot be controlled if they are entirely unknown; instead, we assume there are observed signals that can be used to infer them, but the values themselves are difficult to explicitly enumerate. In addition to confounds introduced by \OPTEXT and \OPTRAITS, \COMMENTTEXT may also contain overt signals, e.g. titles like ``Ma'am'' or ``Sir'', that are predictive of gender, but not indicative of bias. We thus identify 3 factors to account for: \OPTEXT, \OPTRAITS, and overt signals.

\section{Methodology}
\label{sec:methodology}
Our overall methodology centers on creating a classifier that predicts gender of the addressee while controlling for the effects of observed confounding variables (\OPTEXT), latent confounding variables (\OPTRAITS), and overt signals. 
%While there may be other confounds, we focus on \OPTEXT, which we assume is available and therefore observed, and \OPTRAITS, which are difficult to enumerate and therefore latent. 
The input to the prediction model is \COMMENTTEXT, while the output is \OPGENDER, and we aim to identify bias in \COMMENTTEXT.

\subsection{Controlling Observed Confounding Variables through Propensity Matching}
Our primary method for controlling for \OPTEXT is \emph{propensity matching}. Propensity matching was developed to replicate the conditions of randomized trials in causal inference studies \citep{rosenbaum1983central,Rubin85}. In this step, we discard any \COMMENTTEXT training samples whose associated \OPTEXT is heavily affiliated with only one gender. In \Fref{fig:method}, if we assume that only women post ``Do I look ok?'', we would discard all comments posted in reply to the \OPTEXT ``Do I look ok?''. We ultimately seek to balance our data set, so that the set of all \COMMENTTEXT where $\OPGENDER=\mathbf{M}$ has similar associated \OPTEXT as the set of all \COMMENTTEXT where $\OPGENDER=\mathbf{F}$. Thus, we match each \OPTEXT where $\OPGENDER=\mathbf{F}$ with a similar \OPTEXT where $\OPGENDER=\mathbf{M}$ and discard all unmatched data.

Ideally, we would match \OPTEXT values written by men with identical \OPTEXT values written by women, but this is infeasible in practice. Instead, the key insight behind propensity matching is that it is sufficient to match data points based on the probability of the target variable, e.g., the probability that $\OPGENDER=\mathbf{F}$ \citep{rosenbaum1983central,Rubin85}. Thus, the propensity score $e_i$ for a $\COMMENTTEXT_i$ is defined as the probability that $\OPGENDER=\mathbf{F}$, given the confounding variable, $\OPTEXT_i$:
$$ e_i(\COMMENTTEXT_i) = P(\OPGENDER_i = \mathbf{F}|\OPTEXT_i)$$

To balance our data set, we need to ensure that the set of \COMMENTTEXT where $\OPGENDER=\mathbf{M}$ has a similar propensity score distribution as the set of \COMMENTTEXT where $\OPGENDER=\mathbf{F}$. Because propensity scores are dependent on \OPTEXT, all \COMMENTTEXT replied to the same \OPTEXT have the same propensity score. We can then equate $e_i(\COMMENTTEXT_i) = e_i(\OPTEXT_i)$, and focus estimating \OPTEXT scores.

Propensity scores can be estimated by using a classification model that is trained to predict the target attribute $\OPGENDER_i=\mathbf{F}$ from the observed confounding variable $\OPTEXT_i$ \citep{Westreich2010,Lee2010}.
% The general procedure involves first training a prediction model to obtain propensity scores, and then matching samples from the treatment group to samples in the control group based on the scores. Greedy matching (for each treatment sample, the control with the closest propensity score is selected) often performs at least as well as more sophisticated algorithms \citep{Gu1993}.
We use a bidirectional LSTM encoder followed by two feedforward layers with a $\tanh$ activation function and a softmax in the final layer.
Then, we use greedy matching to match each $\OPTEXT_i$ where the true value of $\OPGENDER_i$ is $\mathbf{F}$ with $\OPTEXT_j$ where the true value of $\OPGENDER_j$ is $\mathbf{M}$ and 
$|e_i(\OPTEXT_i) - e_j(\OPTEXT_j)|$
is minimal \citep{Gu1993}. 

We institute a threshold $c$ \citep{Stuart2010}, where we discard $\OPTEXT_i$ if we cannot find a $\OPTEXT_j$ such that $|e_i(\OPTEXT_i) - e_j(\OPTEXT_j)| \le c$. Thus, for example, we would match a post written by a woman that is ``stereotypically female'' (e.g., $e_i$ is large) with a post written by a man that is also ``stereotypically female''  (e.g., $e_j$ is also  large). In \Fref{fig:method}, we match ``Tennis is great'' with ``I love tennis'', and we discard ``Do I look ok?'' as unable to be matched. However, using propensity matching rather than direct matching allows us to match \OPTEXT values that are about different topics, as long as they are equally likely to have been written by a woman.

Finally, our actual model input consists of \COMMENTTEXT, not of \OPTEXT. Once we have matched pairs of \OPTEXT values, we need to ensure that we have an equal number of \COMMENTTEXT values for each \OPTEXT in the pair in order to have a balanced data set.  Then, for each matched $[\OPTEXT_i, \OPTEXT_j]$, we randomly downsample to have an equal number of \COMMENTTEXT values for each \OPTEXT in the pair. In this way, we balance the training set of \COMMENTTEXT in terms of how predictive the confounding variable \OPTEXT is of the target attribute \OPGENDER.

\subsection{Controlling Latent Confounding Variables through Adversarial Training}
\label{sec:method_confound_demotion}

While propensity matching is a desirable way to control for confounding variables because of established literature, matching is only possible for observed variables \citep{Gu1993,Rosenbaum1988}. In our data, while \OPTEXT is observed, \OPTRAITS is not possible to match on (further discussion in \Sref{sec:experimental}).
Instead, we use an adversarial objective drawn from \citet{kumar-etal-2019-topics} to encourage the model to ignore \OPTRAITS.

\paragraph{Confound representation}
While we cannot explicitly enumerate \OPTRAITS, we know that they are associated with the identity of \OP, and we can infer them from \COMMENTTEXT addressed to \OP. We  use associations between \OP and \COMMENTTEXT to derive a feature vector for each $\COMMENTTEXT
_i$ that reflects $\OPTRAITS_i$. The latent confounds to demote are represented as multinomial distributions, derived from log-odds scores \citep{monroe2009}.

For each label $\OP = k$ and each word type $w$ in all \COMMENTTEXT,
we calculate the log-odds score $lo(w, k) \in \mathbf{R}$, where higher scores indicate stronger associations between $k$ and the word. In \Fref{fig:method}, $lo(\texttt{Canada}, \texttt{Person 1})$ would be high, as \COMMENTTEXT values addressed to $\texttt{Person 1}$ often contain the word Canada.
Then, following \citet{kumar-etal-2019-topics}, we define a distribution: for all $k \in \OP$ and an input $\COMMENTTEXT_i$, $ = \langle w_1,\ldots, w_n\rangle$:
\begin{align*}
&p(k | \COMMENTTEXT_i) \propto\\ 
& p(k)p(\COMMENTTEXT_i | k) = p(k)\prod_{i=1}^np(w_i | k) 
\end{align*}
$p(k)$ is estimated from the distribution of $k$ in the training data, i.e., the proportion of \COMMENTTEXT values addressed to $\OP = k$. $p(w_i | k)$ is proportional to $\sigma(lo(w, k))$, where we use the sigmoid function ($\sigma$) to map log-odds scores to the range [0,1] and then normalize them over the vocabulary to obtain valid probabilities. For each input $\COMMENTTEXT_i$, we then obtain a vector whose elements are $p(k | \COMMENTTEXT_i)$ and whose dimensionality is the number of \OP individuals in the training set. We normalize these vectors to obtain multinomial probability distributions which reflect $\COMMENTTEXT_i$'s association with each \OP individual. Thus, when we demote this vector during training, we force the classifier to learn features that are indicative of the group \OPGENDER and not features that are indicative of individual members of this group (e.g., some group members are from Canada). We refer to the confound vector as $t_i$. Justification for the log-odds representation as opposed to alternatives is presented in \citet{kumar-etal-2019-topics}.

\paragraph{Training Procedure}
Our goal is to obtain a model that can predict \OPGENDER, but cannot predict the latent confounds represented by $t_i$. To achieve this, the model is trained in an alternate GAN-like procedure \citep{goodfellow2014generative}.

First, the input $x \in \COMMENTTEXT$ is encoded using an encoder neural network $h(x; \theta_h)$ to obtain a hidden representation $\mathbf{h}_x$. This representation is then passed through two feedforward networks: (1) $c(h(x); \theta_c)$ to predict the label $y \in \{\mathbf{M}, \mathbf{F}\}$; and (2) an adversary network $\mathrm{adv}(h(x); \theta_a)$ to predict the vector representation of the latent confounds.

We train the encoder, so that $\mathbf{h}_x$ does not contain any information predictive of the confound vector, but does contain information predictive of the target attribute.
% We assess if the encoder representation contains predictive information by training the adversary network to predict the confound vector $t_i$ from the encoded input $\mathbf{h}_{x_i}$.
Thus our primary training objective is:
%\begin{equation}
%\label{eq:adv}
%\min_{\mathrm{adv}}\frac{1}{N}\sum_{i=1}^N\mathrm{CE}(\mathrm{adv}(h(x_i)), t_i)
%\end{equation}
\begin{equation*}
\label{eq:class}
%\begin{split}
 \min_{c,h}\frac{1}{N}\sum_{i=1}^N\mathrm{CE}(c(h_{x_i}), y_i) 
 + \mathrm{KL}(\mathrm{adv}(h_{x_i}), \mathbb{U}_K)
%\end{split}
\end{equation*}
where $\mathbb{U}$ represents a uniform distribution, CE represents cross-entropy loss, and KL represents KL-divergence. We refer to \citet{kumar-etal-2019-topics} for the training procedure that alternates minimizing this objective and training the adversary.
 
\subsection{Overt Signals}
Finally, we control for overt signals using word substitutions that replace gendered terms with more neutral language, for example $\textnormal{woman} \rightarrow \langle \textnormal{person} \rangle$  and $\textnormal{man} \rightarrow \langle \textnormal{person} \rangle$. We create a 66-term list of substitutions from existing resources \citep{zhao-etal-2018-gender,bolukbasi2016man} as well as our observations of the data. We also use substitutions to remove the names of addressees from comment, replacing \OP's ``Firstname'' and ``Lastname'' with ``$\langle$name$\rangle$'' in \COMMENTTEXT. We do not attempt to identify nicknames, as the confound demotion method described in \Sref{sec:method_confound_demotion} should already mitigate the influence of individual names, and we perform the substitution as merely an extra precaution.

\section{Experimental Setup}
\label{sec:experimental}

Our primary data is the Facebook subsection of the \href{https://nlp.stanford.edu/robvoigt/rtgender/}{RtGender} corpus \citep{voigt-etal-2018-rtgender}. The data contains two subsections: \textit{Politicians} (400K posts and 13.9M replies addressed to 412 then-current U.S. members of Congress), and \textit{Public Figures} (118K posts and 10.7M replies addressed to 105 famous people such as actresses and tennis players).

%\begin{table}
%    \centering
%    \begin{tabular}{lc|lc}
%    \multicolumn{2}{c}{\textbf{Female Associated}} & \multicolumn{2}{c}{\textbf{Male Associated}}\\
%    women  & -34.00 & Obamacare & 13.39 \\
%    Congresswoman & -29.06 & Iran & 13.23 \\
%    sexual & -23.59 & EPA & 12.55 \\
%    assault & -20.38 & spending & 12.33 \\
%    \end{tabular}
%    \caption{Words from RtGender Politicians \OPTEXT with strong associations with \OPGENDER $\mathbf{M}$ vs.~$\mathbf{F}$.}
%    \label{tab:congress_log_odds}
%\end{table}

We can show that \OPTEXT is a confounding variable by computing log-odds scores between the words in \OPTEXT and \OPGENDER \cite{monroe2009}. In the Politicians data, the most female-associated words are \textit{women}, \textit{Congresswoman}, \textit{sexual}, and \textit{assault}. The most male-associated words are \textit{Obamacare}, \textit{Iran}, \textit{EPA}, and \textit{spending}. It is evident that male and female politicians post about different topics, e.g., female politicians likely post more about sexual assault. A naive model may predict that comments using sexual language are addressed towards women, but increased sexual language may occur because of \OPTEXT, rather than gender bias.

A similar problem occurs with \OPTRAITS, e.g., the corpus has more comments addressed to female tennis players (9 players; 184K comments) than male players (1 player; 29K comments). The model can obtain high accuracy by predicting $\OPGENDER=\mathbf{F}$ for \COMMENTTEXT with the word ``tennis''.  Unlike \OPTEXT, which is observable from the data, we have no way of enumerating every possible value in \OPTRAITS. Even if we could enumerate them, we do not expect propensity matching over \OPTRAITS to work, because we cannot find reasonable matches, e.g., there is only one senior senator from Massachusetts. Additionally, \OPTRAITS can be as fine-grained as names: we cannot find a male senator whom commenters call ``Liz Warren''.

We divide each data set into train, dev, and test sets, enforcing no \OP overlap between subsets. We perform propensity matching and derive the confound vectors to demote using only the training data. We apply word substitutions to all subsets.\footnote{We provide additional details in \Aref{sec:data_statistics}.}

\section{Evaluation}
\label{sec:evaluation}

We train our model to predict \OPGENDER from \COMMENTTEXT, employing propensity matching over \OPTEXT, word substitutions over \COMMENTTEXT, and \OPTRAITS demotion. We focus on evaluating how well our model controls for confounds and whether or not it captures gendered language. Successful demotion of confounds would suggest that our model learns to identify text indicative of gender bias.

\begin{figure}
    \centering
    \includegraphics[width=\columnwidth]{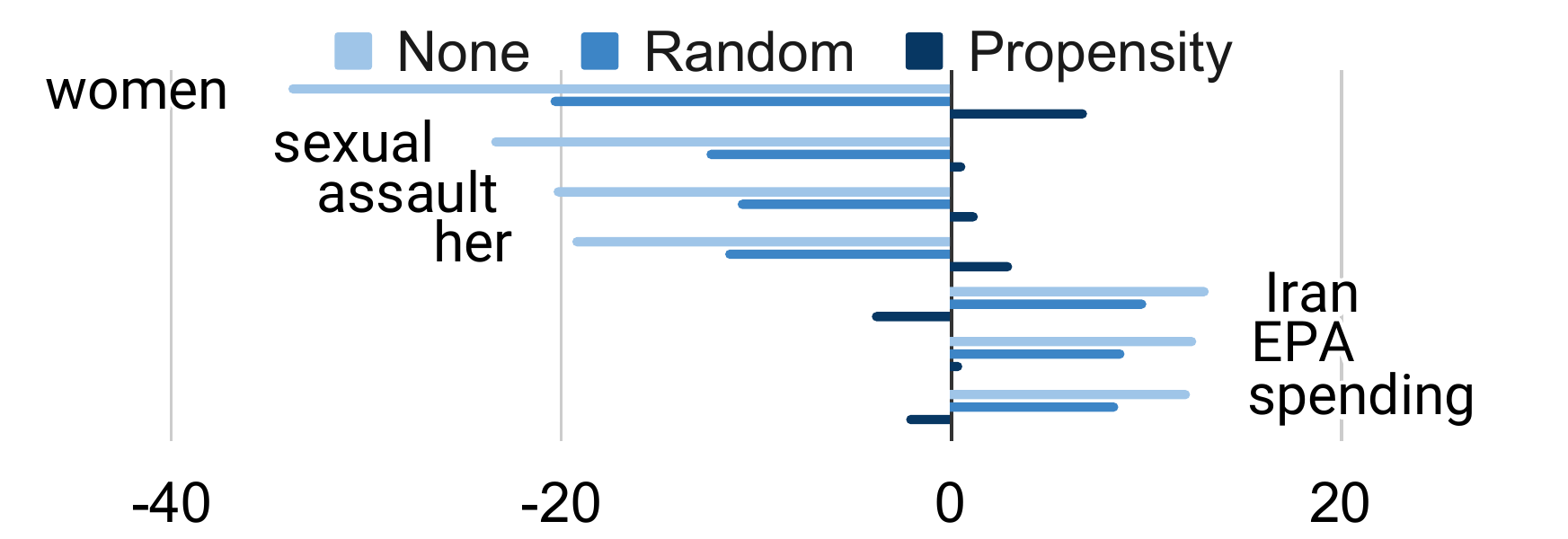}
    \includegraphics[width=\columnwidth]{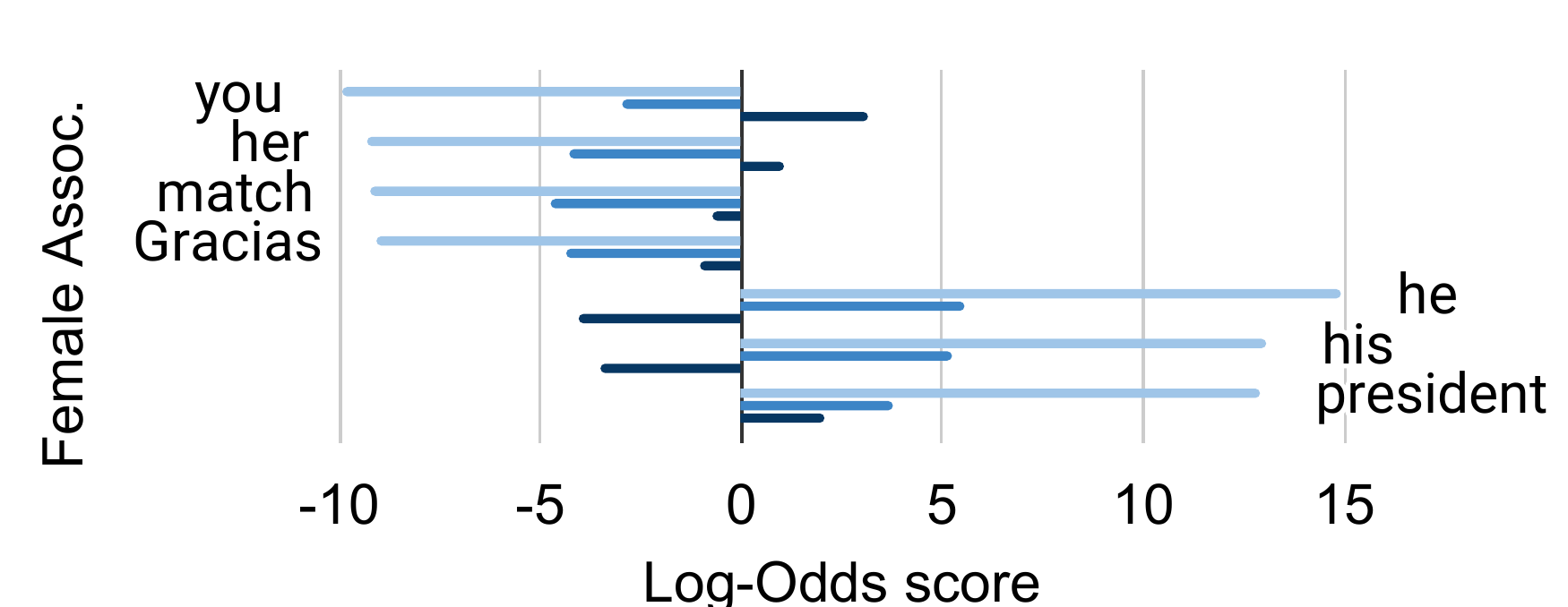}

    \caption{Log-odds scores for most polar words in Politicians (top) and Public Figures (bottom) data, with different matching methods. Propensity matching best reduces polarity.}
    \label{fig:odds}
\end{figure}

\paragraph{Observed Confounding Variable Demotion} In \Fref{fig:odds}, we show log-odds scores, measuring association between \OPTEXT and \OPGENDER in the training set before and after propensity matching. For comparison, we also show scores for a randomly matched data set, in which we balance \OPTEXT to have an equal proportion of $\mathbf{F}$ and $\mathbf{M}$ labels by random sampling (constructed to be the same size as the propensity matched set).
In the Politicians and Public Figures data, propensity matching reduces the magnitude of the most polar words: log-odds scores for the matched data are closer to zero than for the non-matched or randomly matched data.\footnote{Polarities were reduced without producing new ones: in the Politicians data, the magnitude of the 2 most polar words decreased from -34.0 and 17.9 to -7.68 and 8.52, and in the Public Figures data, from -45.5 and 39.3 to -5.29 and 9.43.} Further, propensity matching can even cause the polarity to change direction: words that were originally female-associated (e.g. ``her'') become slightly male-associated. These figures suggest that propensity matching effectively reduces the confounding influence of \OPTEXT.

%NOTE: all the current numbers are from the spreadsheet "Post-ACL Name Substitutions"
\paragraph{Latent Confounding Variable Demotion}
We evaluate how well our model demotes the influence of latent confounding variables over the held-out test sets (\Tref{tab:eval_demo}). We created data splits so that there is no overlap in \OP values between the train and test sets. While there may still be overlap in some latent \OPTRAITS, we expect there to be less overlap in \OPTRAITS between the train and test set than within the train set. Thus, improved performance over the held-out test set would suggest that demotion effectively reduces the influence of the latent confounding variables---the model learns characteristics of comments addressed to women generally rather than characteristics specific to the individual people in the training set. We do not necessarily expect propensity matching to improve performance, as this method reduces the influence of confounding variables that have high overlap between the train and test sets. 

Because the data set is imbalanced (the Politicians test set is $82\% \mathbf{M}$ and the Public Figures test set is $35.9\% \mathbf{M}$), we report F1 and accuracy scores in \Tref{tab:eval_demo}, where \OPGENDER = $\mathbf{F}$ is considered the positive class. As expected, models with demotion perform best on all metrics, with the exception of recall in the Politicians data.\footnote{\Aref{sec:results_extra} reports precision and recall. The discrepencies between F1 and Accuracy are explained by the imbalance in the data set, particularly in the Politicians data set, which is imbalanced in favor of $\mathbf{M}$ while we report metrics assuming $\mathbf{F}$ is the positive class.} We note that in general lack of performance improvement on the test set does not necessarily mean the model is not working, and it could indicate that there is not biased language in the data set. However in this case, since we do observe biased comments in this data (e.g. \Tref{tab:sample_comments}), and we do observe a performance increase, the performance increase suggests that confound demotion improves the model's ability to generalize beyond the individuals in the training set and capture characteristics of language addressed to women in general.

\begin{table}
    \centering
    \begin{tabular}{lccccc}
        & \multicolumn{2}{l}{\textbf{Public Figs}} & & \multicolumn{2}{l}{\textbf{Politicians}}\\
        \cmidrule{2-3} \cmidrule{5-6}
        & \textbf{F1} & \textbf{Acc.} & &  \textbf{F1} & \textbf{Acc.} \\
        \hline
        \hline
        base & 74.9 &	63.8 & & 23.2 & 73.2 \\
        +demotion &	\textbf{76.1} & \textbf{65.1} & & 17.4 & \textbf{77.1} \\
        +match & 65.4 &	56.0 & & 28.5 & 46.7 \\
        +match+dem.  &	68.2 & 59.7 & & \textbf{28.8} & 51.4 \\
    \end{tabular}
    \caption{Evaluation over held-out test sets, where \OPGENDER = $\mathbf{F}$ is considered the positive class. Latent confound demotion improves performance.}
    \label{tab:eval_demo}
\end{table}

\paragraph{Detection of Sexist Comments}
Finally, we evaluate if our model captures gender-biased language by using it to identify gender-based microaggressions, i.e.,``you're too pretty to be a computer scientist!''. This task is notoriously difficult because words like ``pretty'' often register as positive content \citep{breitfeller2019finding,jurgens-etal-2019-just}. Our goal is not to maximize accuracy over microaggression classification, but rather to assess whether or not our model has encoded any indicators of gender bias from the RtGender data set, which would be indicated by better than random performance.

 We use a corpus of self-reported microaggressions.\footnote{Details in \Aref{sec:micro_data}} In the absence of negative examples that contain no microaggressions, we focus on distinguishing gender-tagged microaggressions (704 posts) from other forms of microaggressions, e.g., racism-tagged (900 posts). We train our model on either the Politicians or Public Figures training data sets, and then we test our model on the microaggressions data set. Because most gender-related microaggressions target women, if our model predicts that the reported microaggression was addressed to a woman (e.g. $\OPGENDER=\mathbf{F}$), we assume that the post is a gender-tagged microaggression. Thus, our models are \textit{not trained at all} for identifying gender-tagged microaggressions.

\Tref{tab:eval_micro} shows results from our models and two random baselines. ``Random'' guesses gender-tagged or not with equal probability. ``Class Random'' guesses gender-tagged or not according to true test distributions (56.1\% gender-tagged). All models outperform ``Class random'', and all models with demotion also outperform ``Random''.

Propensity matching improves F1 when training on the Politicians data, but not Public Figures. Several differences could explain this: the Public Figures set is smaller, so propensity matching causes a more substantial size reduction. Also, the Politicians data is more heavily imbalanced, though notably, it is imbalanced in the same direction as the microaggressions data, while the Public Figures data is imbalanced oppositely. Finally, many microaggressions contain references to appearance, which are also common in the Public Figures data. Many comments to people like actresses focus on their looks, especially because they often post photos. However, by controlling for \OPTEXT, propensity matching discards many of these comments. Thus, by demoting a confounding variable, we make the prediction task more difficult. Our goal in confound demotion is not to improve accuracy, but to increase confidence in model outputs.

Nevertheless, the general better-than-random performance of all models is striking, as it suggests strong bias in the underlying training data, which is encoded by our models.

\begin{table}
    \centering
    \begin{tabular}{lccccc}
        & \multicolumn{2}{l}{\textbf{Public Figs}} & & \multicolumn{2}{l}{\textbf{Politicians}}\\
        \cmidrule{2-3} \cmidrule{5-6}
        & \textbf{F1} & \textbf{Acc.} & &  \textbf{F1} & \textbf{Acc.} \\
        \hline
        \hline
        base & 61.3 & 57.3 & & 48.1 & 64.2 \\
        +demotion & \textbf{62.2} & 57.9 & & 53.7 & 61.5 \\
        +match & 38.9 & 55.9 & & 46.9 & 50.7 \\
        +match+dem. & 50.9 & 57.0 & & \textbf{56.9} & 49.9 \\
        \hline 
        \hline
        Random & 46.0 &	49.8 & & - & -\\
        Class Random & 42.1 & 48.3 & & - & - \\
    \end{tabular}
    \caption{Evaluation over the microaggressions data set. Despite not being trained for this task, our models achieve better-than-random performance.} %on correctly labeling gendered microaggressions and surfacing bias.
    \label{tab:eval_micro}
\end{table}

\begin{figure}[ht]
    \centering
    \includegraphics[width=\columnwidth]{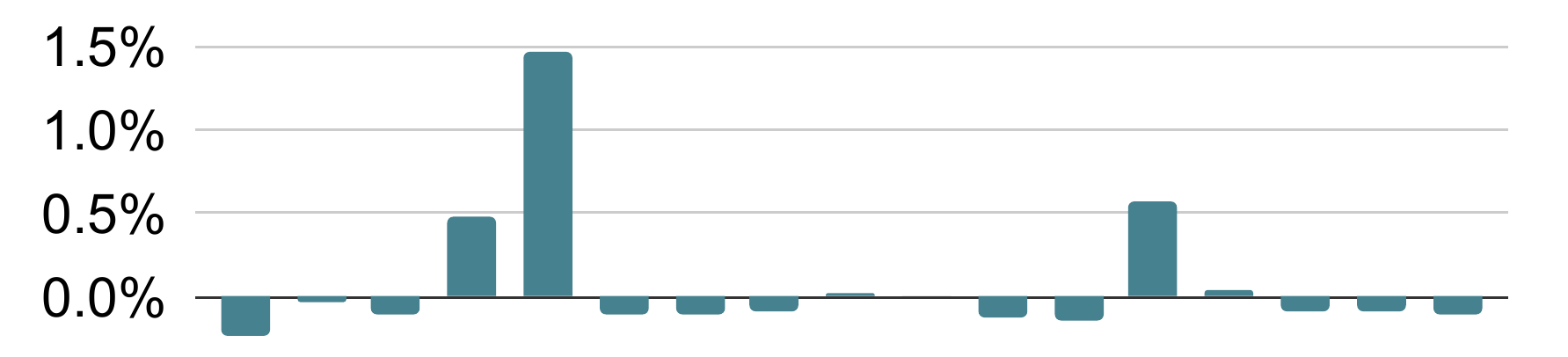}

    \includegraphics[width=\columnwidth]{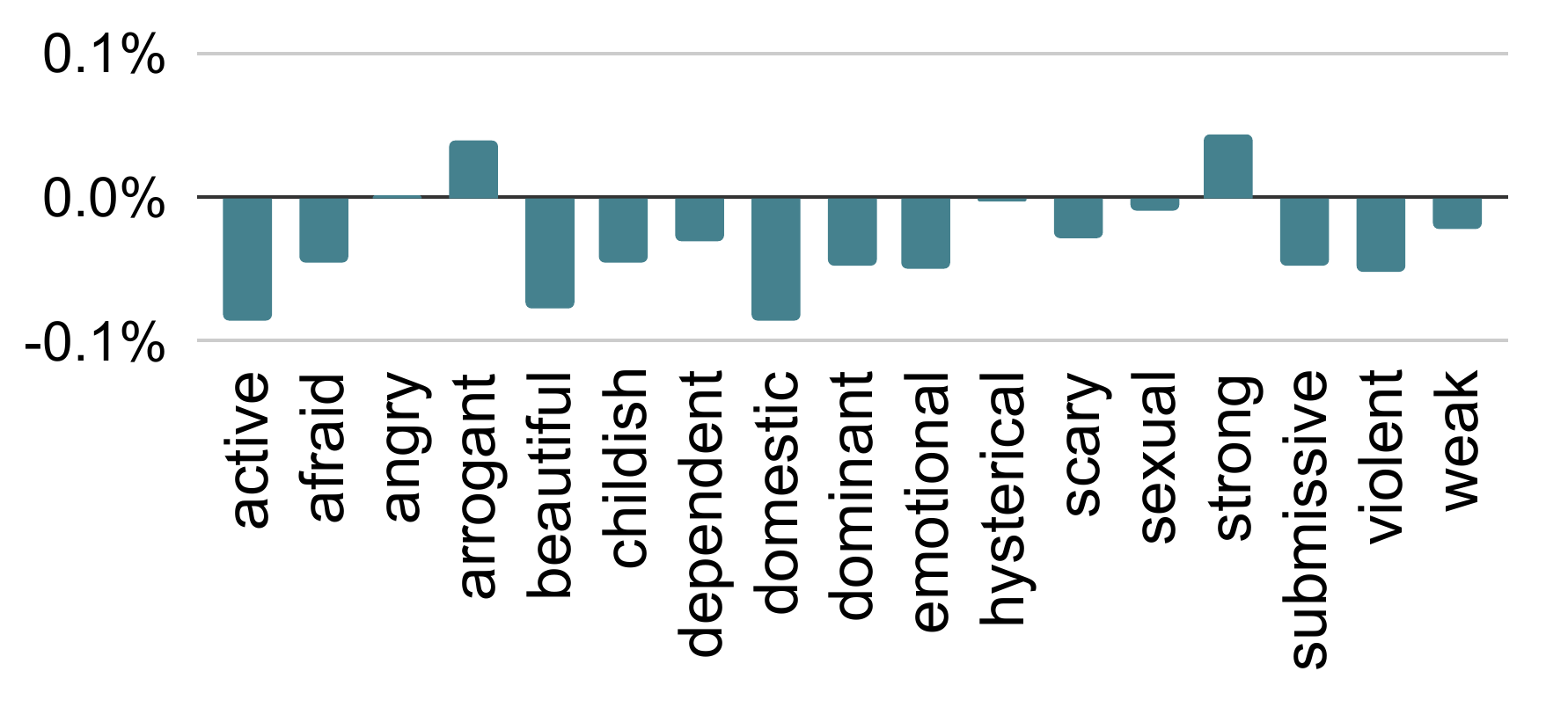}

    \caption{Lexicon differentials between comments with a high likelihood of bias and random samples with $\OPGENDER=\mathbf{W}$ for Public Figures (top) and Politicians (bottom) data. In the Public Figures data, high-likelihood comments are more focused on appearance.}
    \label{fig:stereotype_lexicons}
\end{figure}

\begin{table*}
    \centering
    \begin{tabular}{ll}
%    \hline
%    \textbf{Politicians} & \\
%    \hline
%    \hline
%\OPTEXT & I'd like to wish all of my neighbors and friends who are celebrating today a happy, \\
%        & healthy, and prosperous Lunar New Year! \\
%\COMMENTTEXT & thankyou. thankyou and thankyou dear my love. \\
%\hline

\OPTEXT & 	From reintroducing my legislation to curb sexual assault on college campuses to... \\
%        & Help establish our leadership in the new Senate -- contribute \$5 or more now... \\
\COMMENTTEXT & DINO I hope another real Democrat challenges you next election \\
\hline

% This is from training
\OPTEXT & Donald Trump is the President, not our ruler...Speak up!  Call the White House... \\
% \OPTEXT & Donald Trump is the President, not our ruler. The ugliness that is spilling out of the \\
%        & White House is awful, but love trumps hate. Stand up. Speak up!  Call the White House... \\
        % Call Speaker Ryan. Write a letter to the editor. Call a talk show. Organize. Fired up, ready to go? Yes! \\
\COMMENTTEXT & $\langle$name$\rangle$ Shea-Porter, I did not vote for you and have no clue why anyone should have. \\
            & You do not belong in politics	 \\

% \hline
% \textbf{Public Figures} & \\
\hline
\hline

% \OPTEXT & Celebrating a halfway done party with the cast and crew of ``Fun Size". Such an \\
%        & awesome group of people, I feel so lucky. \\
% \COMMENTTEXT & tori, receive a kiss fm me, $\langle$unk$\rangle$ \\
% \hline
% \OPTEXT & On my way the VMA's! It's going to be a great show. \\
% & Who do you guys think I'm going to be presenting with? Let's see if you can guess ;\{) \\
% \COMMENTTEXT & I $\heartsuit \heartsuit \heartsuit \heartsuit \heartsuit \heartsuit \heartsuit \heartsuit \heartsuit$ your dress $\langle$unk$\rangle$ much!!!! it takes a real beauty to rock that \\
% & dress, and you were the only one pretty enough to wear it!!! \\
%\hline
% These ones are from training
\OPTEXT & I am wondering about the guy who actually cried over spilt milk? He must have had... \\
\COMMENTTEXT & Total tangent I know but, you're gorgeous. \\
\hline
\OPTEXT & Bob and I join Bill Hemmer on America's Newsroom to discuss whether or not... \\
\COMMENTTEXT & I like Bob, but you're hot, so kick $\langle$theirs$\rangle$ butt. \\

% This one is from training

    \end{tabular}
    \caption{Example comments surfaced by our model from Politicians (top) and Public Figures (bottom) data sets.}
    \label{tab:sample_comments}
\end{table*}

\section{Analysis of Encoded Bias}
\label{sec:analysis}

Finally, we analyze what type of bias our model learns: (1) we identify words that most impact model confidence; (2) we compare posts surfaced by our model with prior work on stereotypes; (3) we show example posts surfaced by our model. Throughout this section, we use \textit{prediction score} to refer to the output of the final softmax layer of the prediction model, which we take as an estimate of model confidence. We generally focus on \COMMENTTEXT for which our model predicts $\OPGENDER=\mathbf{F}$ with a high prediction score. These are the posts our model identifies as likely to contain bias against women: despite the matching and demotion methods, the model still predicts $\OPGENDER=\mathbf{F}$ with high confidence.

\paragraph{Influential words} We identify words that strongly influence the model's decisions by masking out words from comments in the test set and examining the impact on prediction score. For each data set, we take the 500 comments from the test set for which the model predicts $\OPGENDER=\mathbf{F}$ with maximal prediction scores. We then generate masked posts: for every word $w$ in the post, we generate a version of the post that omits $w$. We run these masked posts through our gender-prediction model and compare the prediction scores where $w$ is omitted and where $w$ is not omitted, averaging across all occurrences of $w$ in the 500 posts. We then examine the set of $w$ words with the highest differential in prediction score - these are words that, when omitted, cause the model to less associate $\OPGENDER$ with $\mathbf{F}$.

In the Public Figures data, the most influential words are appearance-driven and sexualized: \textit{beautiful}, \textit{bellissima}, \textit{amore},  \textit{amo}, \textit{love}, \textit{linda}, \textit{sexo}. In contrast, influential words in the Politicians data are more mixed. Words include references to strength and competence, e.g., \textit{force}, \textit{situation}, as well as traditionally domestic terms, e.g., $\langle$\textit{spouse}$\rangle$\footnote{``$\langle \rangle$'' indicate overt terms substituted out. ``$\langle$spouse$\rangle$'' replaced ``husband'', ``husbands'', ``wife'', and ``wives''.}, \textit{family}, \textit{love}. When we repeat this process using the 500 highest-confidence posts from the training set instead of the test set, we find similar results. Influential words in the Public Figures training data primarily refer to appearance, while ones in the Politicians training data include terms like \textit{DINO}.\footnote{``Democrat in Name Only'' a political insult} However, influential words from the training data also includes some correlative terms, like names of states, that we would expect the latent confound demotion to de-emphasize. While \Sref{sec:evaluation} suggests that our model successfully reduces the influence of confounding variables, more work is needed to eliminate them entirely.

\paragraph{Comparison to stereotype lexicons}
In order to better understand these trends, we draw from prior work on stereotype detection \citep{fast2016shirtless}. We take the set of test comments for which our model predicts $\OPGENDER=\mathbf{F}$ with a high prediction score ($\ge0.99$ for Public Figures; $\ge0.95$ for Politicians). Then, we compute the difference in frequency of words from a stereotype lexicon \citep{fast2016shirtless} in this high-confidence prediction set with their frequency in a random sample of the same number of comments where the true value of $\OPGENDER=\mathbf{F}$.\footnote{We ignore non-English comments and lemmatize the comment text and lexicons. We randomly sample twice and average frequencies between samples. Lexicon counts are normalized by total number of words in the sample.} 

\Fref{fig:stereotype_lexicons} reports results, which reflect the same trends observed in the influential words. In the Public Figures data, the lexicons that overlap the most with the high-bias posts are ``Beautiful'', ``Arrogant'', and ``Sexual'', which suggests that bias in these comments focuses on appearance and sexualization. In contrast, bias in comments directed towards politicians are less focused, and differences between the high-confidence prediction posts and the random sample are smaller. The two most prominent lexicons are ``Arrogant'' (primarily driven by lexicon words \textit{special}, \textit{proud}) and ``Strong''. Notably, we do not account for negation of lexicon words. A narrative of power is reflected in comments surfaced by our model: ``you \& Nikki Haley lost my vote on the flag issue \textit{your both weak}''. We provide more examples in \Tref{tab:sample_comments}.

Because the stereotype lexicons are small and scores can be dominated by a few words, we also compare LIWC scores \citep{pennebaker2001linguistic}. While most LIWC categories are too broad to align with well-known stereotypes, results are consistent with \Fref{fig:stereotype_lexicons}; for Public Figures, the high-bias data scores higher than the random sample for the ``Sexual'' (0.32 vs.~0.10) and ``Body'' (0.70 vs.~0.56). For Politicians, the high-bias comments score lower than the random sample in the ``Drives'' (8.76 vs.~9.71), which encompasses Affiliation, Achievement, Power, Reward, and Risk focus.

The difficulty in evaluating our model against existing lexicons as well as the differences between the two data sets motivates our goal in learning to detect bias automatically. Bias can differ in different contexts, making it difficult to crowdsource through annotations or define through lexicons.

\paragraph{Examples}

\Tref{tab:sample_comments} shows training and test examples surfaced by our model. We identify them by selecting posts where \OPTEXT is not strongly gendered (propensity score model described in \Sref{sec:methodology} outputs a prediction score $<0.6$), but where \COMMENTTEXT is strongly gendered ($>0.9$ prediction score).
%\footnote{We also discard all \OPTEXT posts that contained photo or video attachments, leaving text-only posts.}
While posts from the Politicians data are diverse, posts from the Public Figures data focus on appearance. These comments reflect the broader trends shown in the influential words and in \Fref{fig:stereotype_lexicons}.

\section{Related Work}

Our work differs from prior work on bias detection in NLP in that we infer bias from data in an unsupervised way, whereas prior work relies on crowd-sourced annotations \citep{fast2016shirtless,bolukbasi2016man,wang-potts-2019-talkdown,sap2019social}. This work typically focuses on specific types of bias, such as condescension \citep{wang-potts-2019-talkdown} or microaggressions \citep{breitfeller2019finding} and involves carefully constructed annotations schemes that are difficult to generalize to other data sets or types of bias. In contrast, our unsupervised approach is not limited to any particular domain and does not rely on human annotations, which can be subjective.

Less-supervised approaches focus on corpus-level analyses, such as associations between gendered terms and occupational stereotypes \citep{wagner2015s,bolukbasi2016man,fu2016tie,joseph2017girls,nakandala2017gendered,friedman-etal-2019-relating,chaloner-maldonado-2019-measuring,hoyle-etal-2019-unsupervised}. Methodologies for identifying gender-related differences in text have varied, including word-embedding similarity \citep{bolukbasi2016man}, language model perplexity \citep{fu2016tie}, and predictive words identified by logistic regression \citep{nakandala2017gendered}. These metrics are meaningful over a corpus-level, but are often difficult to interpret over short text spans. Additionally, none of these methods focus on controlling for confounds. 

While matching is a well-established method for controlling for confounding variables in causality literature \citep{rosenbaum1983central,Rubin85,Stuart2010}, considerably less work has drawn this methodology into NLP. Most work takes one of two approaches. In the first scenario, text maybe be a confounding variable that needs to be controlled in order to measure the effect of a non-text variable \citep{roberts2018adjusting,veitch2019using}. For example, \citet{roberts2018adjusting} examine whether or not papers written by male authors are cited more than ones by female authors, while controlling for the content of the paper. \citet{roberts2018adjusting} also offer a specific method for matching text, which relies on the output of a topic model. In this work, we use the output of an LSTM, which is generally more appropriate for short text, does not make the simplifying BOW assumption, and scales well to large data sets.

In the second scenario, it may be desirable to control for non-text confounds before analyzing text. \citet{chandrasekharan2017you} use matching to identify similar users on Reddit before comparing the content that they post. Our work requires both of these perspectives, as the variable we control for (\OPTEXT) and the outcome we analyze (\COMMENTTEXT) are both text. \citet{egami2018make} do consider a similar setting where text is both an outcome and a confound. While their goals differ greatly from ours, our framework is generally consistent with their recommendations.
\citet{keith2020text} provide a more complete overview of using text to reduce the influence of confounding variables.

\section{Limitations and Future Work}
While our work serves as an initial approach toward unsupervised detection of comment-level gender bias, we identify several limitations and areas for future work. We first focus on limitations within our proposed framework. First, while our results in \Sref{sec:evaluation} suggest that adversarial training does help reduce the influence of latent confounding variables, the analysis in \Sref{sec:analysis} suggests that there is scope for improvement. Furthermore, while we focus on some confounds in the data, there may be additional ones that our model does not account for, such as the impact of videos, photos, or links shared with \OPTEXT. Similarly, while our model uses \OPTEXT for propensity matching in the training data, thus encouraging the model to encode indicators of bias, a model to classify comments as biased or unbiased should also incorporate \OPTEXT when assessing test data.  Additionally, we assume that all comments are directly addressed to \OP, but some comments may be addressed to other commenters.
Finally, our assumption that human judgements are not reliable for this task makes evaluation difficult, and this task would benefit from the development of additional evaluation metrics.

There are additional avenues for future work beyond our proposed framework. Notably, we focus on the perspective of \OP and examine what bias social media users may be exposed to, i.e. what comments men and women might expect to receive in response to their posts. We do not examine why comments addressed toward men and women may differ, whether because the same commenters write different comments to men and women, or because men and women attract comments from different types of people. This perspective would require controlling for traits of the commenter, such as gender, age, and occupation. Nevertheless, our work stands without this perspective: biased comments are harmful to the recipient, regardless of who wrote them.

\section{Conclusions}
Bias detection is useful for fostering civil communication on social media, as it allows recipients to screen out biased comments.
Further, our intention is to detect implicit bias that people may not know they have - revealing these biases to social media users could proactively prevent them from posting unintentionally biased comments.
Detecting and analyzing bias is a first step towards mitigating it, and we hope our work will encourage future work in this area.

\section*{Acknowledgements}
We would like to thank reviewers and area chairs, as well as Vidhisha Balachandran, Amanda Coston,  Xiaochuang Han, Sachin Kumar, Artidoro Pagnoni, Chan Young Park, and Shuly Wintner for their helpful feedback on this work.
This material is based upon work supported by the NSF Graduate Research Fellowship Program under Grant No.~DGE1745016, the Google PhD Fellowship program, NSF grants IIS1812327 and SES1926043, an Okawa Grant, and the Public Interest Technology University Network Grant No.~NVF-PITU-Carnegie Mellon University-Subgrant-009246-2019-10-01.
We would also like to thank Amazon for providing GPU credits.
Any opinions, findings, and conclusions or recommendations expressed in this material are those of the authors and do not necessarily reflect the views of the NSF.

\bibliography{new}
\bibliographystyle{acl_natbib}

\newpage
\appendix
\setcounter{page}{1}
\section{Data and Model Implementation Details}
\label{sec:data_statistics}

\begin{table}[ht]
    \centering
\begin{tabular}{lcc}
	& Politicians &	Pub. Figures \\
Raw train size	& 6.9M	& 4.2M \\
Test/Dev size	& 2.3M/2.5M	& 1.9M/0.55M \\
\hline
\% $\mathbf{M}$ in train	& 71.3\%	& 33.9\% \\
Matched train  size	& 256K	& 77K \\
Raw dem. dim.	& 240	& 63 \\
Matched dem. dim.	& 239	& 60 \\

\end{tabular}
    \caption{Data Statistics. ``Matched train size'' refers to the size of the training set after propensity matching, and ``dem. dim.'' refers to the size of the latent confound vector that is demoted during training.}
    \label{tab:data_stats}
\end{table}

All data is lowercased and tokenized, and we discard data points with fewer than 4 tokens. \Tref{tab:data_stats} reports details of our data set after preprocessing.

 For the primary prediction models, we use the same architectures as \citet{kumar-etal-2019-topics}, including training multiple (2) adversaries. We perform minimal hyper-parameter tuning, primarily using the same parameters as \citet{kumar-etal-2019-topics}, with the exception of the learning rate, which we changed slightly to decrease fluctuations in validation accuracy, and the number of training epochs for each phase of the model, which we increased or decreased as needed based on how long the validation accuracy improved for. These changes were determined by manual tuning over $<10$ trials. For the propensity score model, we use a learning rate of 1e-3. For all other models  we use a learning rate of 1e-4. For the models without confound demotion, we train for 5 epochs. For the models with confound demotion, we train the classifier for 3 epochs, the adversary for 10 epochs, and we repeat the alternating cycle for 3 epochs. For all models, we choose the best model as measured by \OPGENDER classification accuracy over the validation set. Each model was trained using 1 GPU. The models without latent confound demotion and the propensity score estimation model have 4.2M parameters each. The adversary in the latent confound demotion models adds an additional 61.7K parameters to the Politicians model and 16.2K parameters to the Public Figures model.
 
 \section{Additional Evaluation Metrics}
 \label{sec:results_extra}
 
 \Tref{tab:eval_demo_extra} provides the same results as \Tref{tab:eval_demo}, with the addition of precision and recall scores. \Tref{tab:eval_val} shows results for the same experiments as \Tref{tab:eval_demo_extra}, but provides metrics over the validation sets instead of the test sets. \Tref{tab:eval_micro_extra} extends \Tref{tab:eval_micro} by additionally showing precision and recall scores.
 
 \begin{table}[!ht]
    \centering
    \begin{tabular}{lcccc}
            \hline 
        & \textbf{Prec.} & \textbf{Rec.} & \textbf{F1} & \textbf{Acc.} \\
        \hline 
        \multicolumn{4}{l}{\textbf{Public Figures}} \\
        base & 67.3 & 84.5 & 74.9 &	63.8 \\
        +demotion & 67.8 & \textbf{86.7} &	\textbf{76.1} & \textbf{65.1} \\
        +match & 65.9 & 65.0 & 65.4 &	56.0 \\
        +match+demotion & \textbf{69.0} & 67.5 &	68.2 & 59.7 \\
    \hline 
     \multicolumn{4}{l}{\textbf{Politicians}} \\
    base	& 24.0 & 22.4 & 23.2 & 73.2 \\
    +demotion	& \textbf{24.8} & 13.4 & 17.4 & \textbf{77.1} \\
    +match	& 18.8 & \textbf{58.8} & 28.5 & 46.7 \\
    +match+demotion	& 19.5 & 54.4 & \textbf{28.8} & 51.4 \\
    \end{tabular}
    \caption{Evaluation over held-out test sets, where \OPGENDER = $\mathbf{F}$ is considered the positive class, extending \Tref{tab:eval_demo} by showing precision and recall.}
    \label{tab:eval_demo_extra}
\end{table}

 \begin{table}[!ht]
    \centering
    \begin{tabular}{lcccc}
            \hline 
        & \textbf{Prec.} & \textbf{Rec.} & \textbf{F1} & \textbf{Acc.} \\
        \hline 
        \multicolumn{4}{l}{\textbf{Public Figures}} \\
        base & 61.4	 & 76.8 & 68.3 & 57.6 \\
        +demotion & 61.6 & 79.5 & 69.4 & 58.5 \\
        +match & 61.0 &	61.2 &	61.1 & 53.8 \\
        +match+demotion & 64.1 & 55.6  & 59.5 &	55.2 \\
    \hline 
     \multicolumn{4}{l}{\textbf{Politicians}} \\
    base & 24.0 & 20.1 & 21.9 &	68.5 \\
    +demotion & 26.3 & 13.0 & 17.4 & 72.9 \\	
    +match	& 21.6 & 55.9 &	31.2 & 45.9 \\
    +match+demotion	& 22.8 & 53.3 &	31.9 & 50.2 \\
    \end{tabular}
    \caption{Evaluation over validation sets, where \OPGENDER = $\mathbf{F}$ is considered the positive class, provided for reproducibility.}
    \label{tab:eval_val}
\end{table}

\begin{table}[!t]
   \centering
    \begin{tabular}{lcccc}
        \hline 
        & \textbf{Prec.} & \textbf{Rec.} & \textbf{F1} & \textbf{Acc.} \\
        \hline 
        \multicolumn{4}{l}{\textbf{Public Figures Training Data}} \\
        base & 50.9	& 77.0 & 61.3 & 57.3 \\
        +demotion & 51.3 & 79.0 & \textbf{62.2} & 57.9 \\
        +match & 49.7 & 32.0 & 38.9 & 55.9 \\
        +match+demotion & 51.0 & 50.7 & 50.9 & 57.0 \\
        \hline 
        \multicolumn{4}{l}{\textbf{Politicians Training Data}} \\
        base & 66.0 & 37.8 & 48.1 & 64.2 \\
        +demotion & 59.6 & 48.9 & 53.7 & 61.5 \\
        +match & 44.5 & 49.6 & 46.9 & 50.7 \\
        +match+demotion & 45.7 & 75.3 & \textbf{56.9} & 49.9 \\
        \hline
        Random & 43.5 &	48.7 &	46.0 &	49.8 \\
        Class Random & 41.4 &	42.9 &	42.1 &	48.3 \\
    \end{tabular}
    \caption{Evaluation over the microaggressions data set, extending \Tref{tab:eval_micro} by showing precision and recall.}
    \label{tab:eval_micro_extra}
\end{table}

\section{Microaggressions Data Set}
\label{sec:micro_data}
 
The dataset of microagressions is taken from \citet{breitfeller2019finding}, who collected the corpus from \url{www.microaggressions.com}. On this website, posters describe a microaggression that they experienced. They can using quotes, transcripts, or narrative text to describe the experience, and these posts are tagged with type of bias expressed, such as ``gender'', ``ableism'', ``race'', etc. We discard all posts that contain only narrative text, since it is not 2\textsuperscript{nd} person perspective and thus very different than our training data, which leaves 1,604 posts for analysis.

\end{document}